\documentclass[preprint,1p]{elsarticle}
\pdfoutput=1
\usepackage{pdflscape}
\usepackage{booktabs}
\usepackage{multirow}
\usepackage{subfigure}
\usepackage{url}

\usepackage{amsmath,amssymb,amsfonts}
\usepackage{mathtools}
\usepackage{float}
\usepackage{graphicx}
\usepackage{gensymb}
\usepackage{makecell}

\usepackage{amsthm}
\usepackage{epsfig}

\newcommand{\figref}[1]{Fig. \ref{#1}}
\newcommand{\tabref}[1]{Table \ref{#1}}
\newcommand{\secref}[1]{Section \ref{#1}}

\journal{arXiv}

\begin{document}

\begin{frontmatter}

\title{G-Rep: Gaussian Representation for Arbitrary-Oriented Object Detection}

\author[1]{Liping Hou}
\ead{houliping17@mails.ucas.ac.cn}

\author[1,2]{Ke Lu}
\ead{luk@ucas.ac.cn}

\author[2]{Xue Yang}
\ead{yangxue-2019-sjtu@sjtu.edu.cn}

\author[1]{Yuqiu Li}
\ead{liyuqiu20@mails.ucas.ac.cn}

\author[1]{Jian Xue\corref{cor}}
\ead{xuejian@ucas.ac.cn}

\cortext[cor]{Corresponding author}
\address[1]{School of Engineering Science, University of Chinese Academy of Sciences, Beijing 100049, China}
\address[2]{Peng Cheng Laboratory, Shenzhen 518055, China}
\address[3]{Department of Computer Science and Engineering, Shanghai Jiao Tong University, Shanghai 200240, China}

\begin{abstract}
Typical representations for arbitrary-oriented object detection tasks include oriented bounding box (OBB), quadrilateral bounding box (QBB), and point set (PointSet). Each representation encounters problems that correspond to its characteristics, such as the boundary discontinuity, square-like problem, representation ambiguity, and isolated points, which lead to inaccurate detection. Although many effective strategies have been proposed for various representations, there is still no unified solution. Current detection methods based on Gaussian modeling have demonstrated the possibility of breaking this dilemma; however, they remain limited to OBB. To go further, in this paper, we propose a unified Gaussian representation called G-Rep to construct Gaussian distributions for OBB, QBB, and PointSet, which achieves a unified solution to various representations and problems. Specifically, PointSet or QBB-based object representations are converted into Gaussian distributions, and their parameters are optimized using the maximum likelihood estimation algorithm. Then, three optional Gaussian metrics are explored to optimize the regression loss of the detector because of their excellent parameter optimization mechanisms. Furthermore, we also use Gaussian metrics for sampling to align label assignment and regression loss. Experimental results on several public available datasets, such as DOTA, HRSC2016, UCAS-AOD, and ICDAR2015, show the excellent performance of the proposed method for arbitrary-oriented object detection.
\end{abstract}


\end{frontmatter}

\section{Introduction} \label{section:introduction}
With the development of deep convolutional neural network (CNN), the object detection, especially arbitrary-orientation object detection
\cite{azimi2018towards,ding2018learning,yang2019scrdet,yang2021r3det,yang2021rethinking}, has been developed rapidly and a variety of methods have been proposed.

\begin{figure}[htb]
\centering
\includegraphics[width=0.65\linewidth]{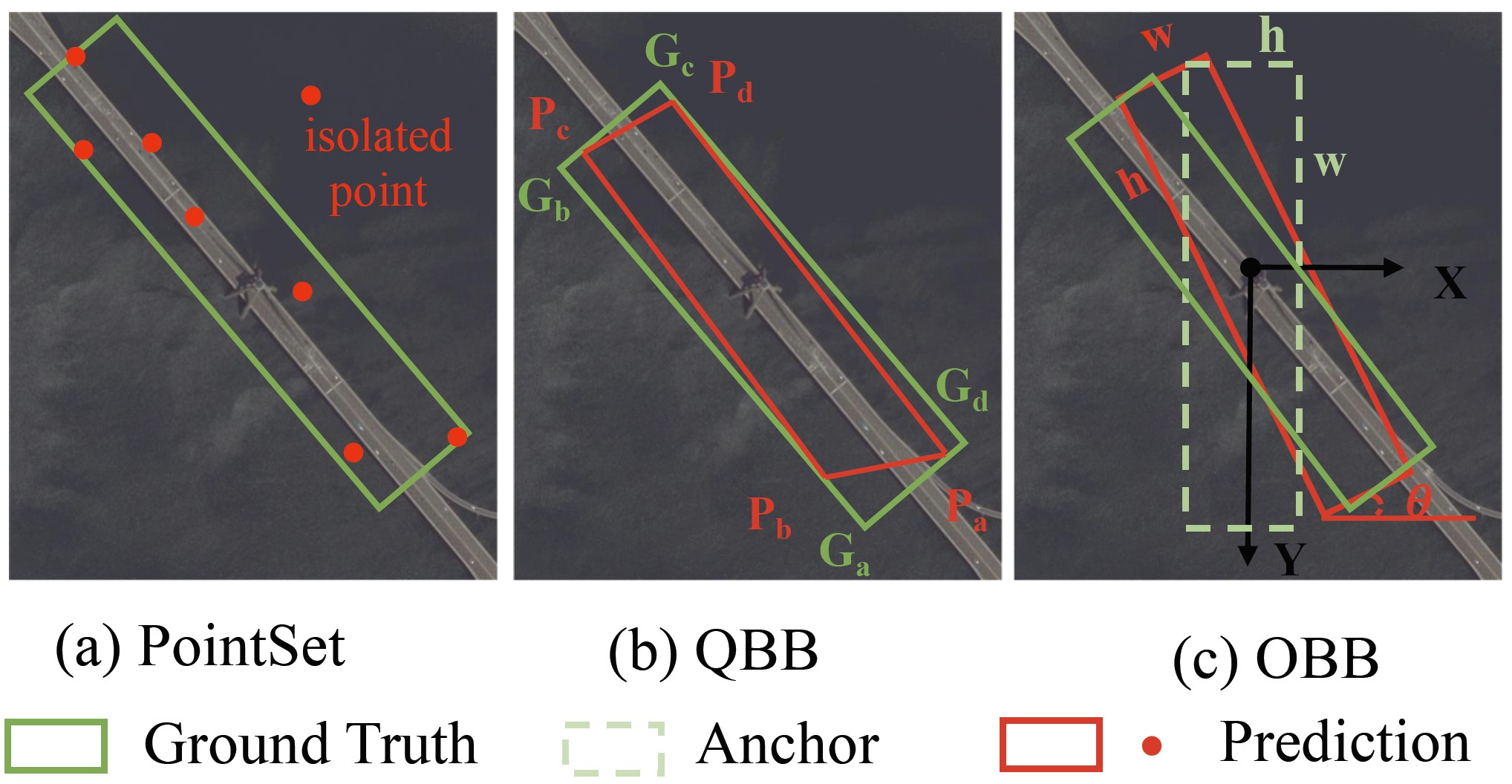}
\caption{Illustrations of different problems for different representations: (a) Dissociation of PointSet. (b) Representation ambiguity of QBB. (c) Boundary discontinuity of OBB.} \label{fig:questions}
\end{figure}

Popular representations of the arbitrary-oriented object are divided into oriented bounding box (OBB) \cite{yang2018automatic}, quadrilateral bounding box (QBB) \cite{ming2021optimization}, and point set (PointSet) \cite{guo2021beyond}.
Each representation encounters intrinsic issues because of the properties of its definitions, which are summarized as follows and illustrated in \figref{fig:questions}.
\begin{enumerate}
    \item  PointSet uses several individual points to represent the overall arbitrary-oriented object, and the independent optimization between the points makes the trained detector very sensitive to isolated points, particularly for objects with large aspect ratios because a slight deviation causes a sharp drop in the intersection over union (IoU) value. As shown in \figref{fig:questions}(a), although most of the points are predicted correctly, an outlier makes the final prediction fail. Therefore, the joint optimization loss (e.g., IoU loss \cite{yu2016unitbox,rezatofighi2019generalized,zheng2020distance}) based on the point set is more popular than the independent optimization loss (e.g., $L_{n}$ loss).
    
    \item  As a special case of PointSet, QBB is defined as the four corners of a quadrilateral bounding box. In addition to the inherent problems of PointSet described above, QBB also suffers from the representation ambiguity problem \cite{ming2021optimization}.
    Quadrilateral detection often sorts the points first (as shown in \figref{fig:questions}(b), represented by the green box) to facilitate the point matching between the ground-truth and prediction bounding boxes to calculate the final loss.
    Although the red prediction box in \figref{fig:questions}(b) does not satisfy the sorting rule and gets a large loss value accordingly using the $L_{n}$ loss, this prediction is correct according to the IoU-based evaluation metric.

    \item  OBB is the most popular choice for oriented object representation because of its simplicity and intuitiveness. However, the boundary discontinuity and square-like problem are obstacles to high-precision locating, as detailed in \cite{yang2020arbitrary,yang2021dense,yang2021rethinking,yang2021learning}. \figref{fig:questions}(c) illustrates the boundary problem of OBB representation, considering the OpenCV acute angle definition ($\theta \in [-\pi/2, 0)$) as an example \cite{yang2018automatic}. The height ($h$) and width ($w$) of the box swap at the angle boundary, resulting in a sudden change in the loss value, coupled with the periodicity of the angle, which makes regression difficult.
\end{enumerate}

To improve detection performance, many researchers have proposed solutions for some issues, which mainly include the boundary discontinuity \cite{yang2019scrdet,yang2020arbitrary}, square-like problem \cite{yang2021dense, qian2021learning}, representation ambiguity \cite{ming2021optimization}, and isolated points. 
For instance, in previous works \cite{yang2019scrdet,yang2020arbitrary}, the authors aimed to solve the boundary discontinuity in OBB. DCL \cite{yang2021dense} dynamically adjusts the periodicity of loss weights through the aspect ratio to alleviate the square-like problem in OBB. 
RSDet \cite{qian2021learning} uses the modulated loss to smooth the boundary loss jump in both OBB and QBB.
RIDet \cite{ming2021optimization} uses the Hungarian matching algorithm to eliminate the representation ambiguity caused by the ordering of points in QBB.
However, there is still no concise and unified solution to these problems because different solutions require different model settings.

\begin{figure}[htb]
\centering
\includegraphics[width=0.99\linewidth]{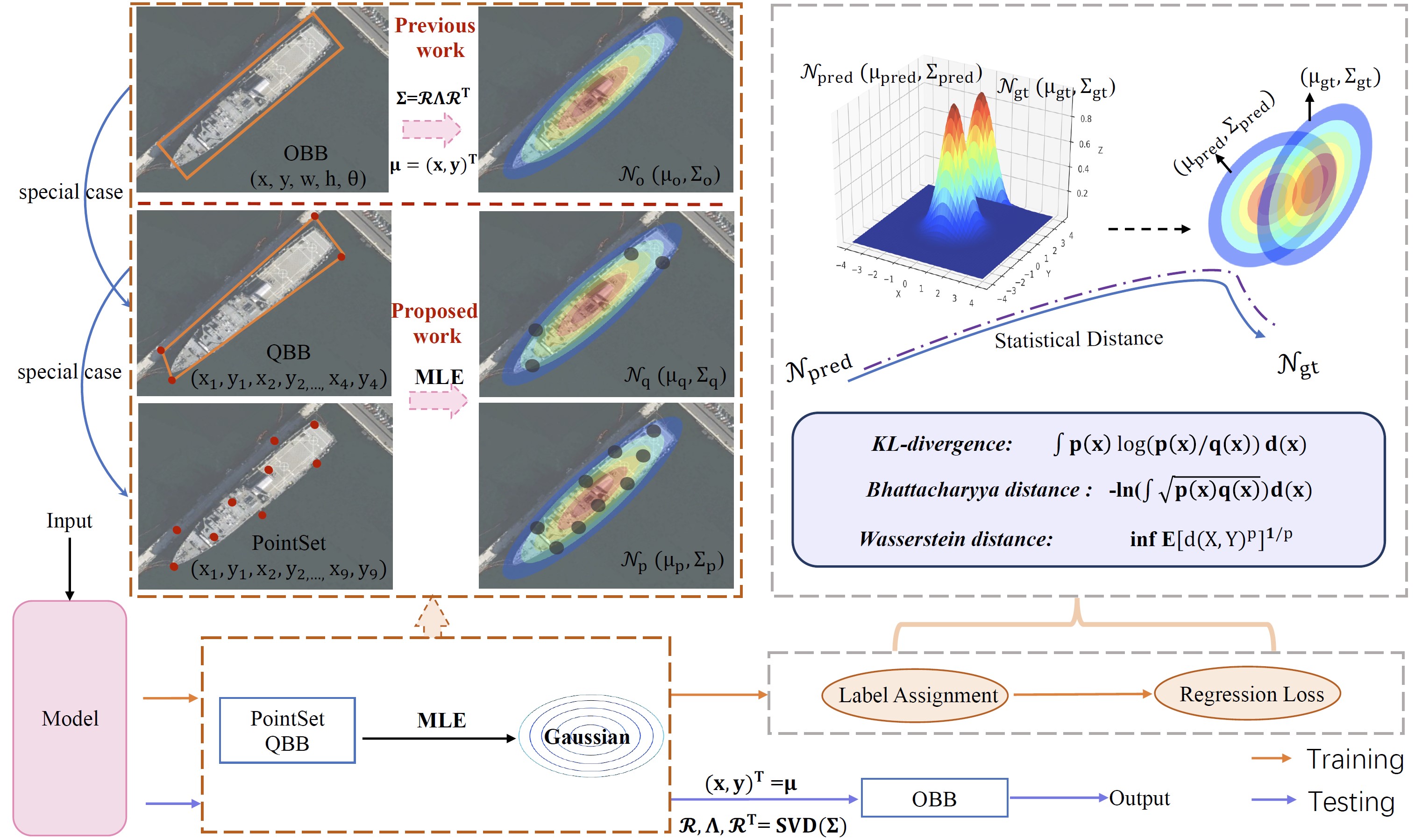}
\caption{Overview of the main contributions of this paper. Gaussian distributions of QBB and PointSet are constructed, and label assignment strategies and regression losses are designed in an alignment manner on the basis of statistical distances.} \label{fig:overall}
\end{figure}

Recent methods GWD \cite{yang2021rethinking} and KLD \cite{yang2021learning} have broken the paradigm of existing regular regression frameworks from a unified regression loss perspective.
Specifically, OBB is mapped to the Gaussian distribution using a matrix transformation, as shown in the upper left of \figref{fig:overall}, and then robust Gaussian regression losses are designed.
Although promising results have been achieved, the limitations of the OBB representation makes them not truly unified solutions in terms of representation.
Additionally, although several distances for the Gaussian distribution have been explored and devised as the regression losses, the metric for dividing positive and negative samples in label assignment (i.e, sample selection) has not been changed accordingly and IoU is still used. 

In this paper, we aim to develop a fully unified solution to the problems that result from various representations.
Specifically, the PointSet representation and its special cases, QBB, are converted into the Gaussian distribution, and the parameters of Gaussian distribution are evaluated using the maximum likelihood estimation (MLE) algorithm \cite{1977Maximum}.
Furthermore, three evaluation metrics are explored for computing the similarity of two Gaussian distributions, and Gaussian regression losses based on these metrics are designed.
Accordingly, the selection of positive and negative samples are modified using the Gaussian metric by designing fixed and dynamic label assignment strategies. The entire pipeline is shown in \figref{fig:overall}.
The highlights of this paper are as follows.
\begin{itemize}
\item To uniformly solve the different problems introduced by different representations (OBB, QBB, and PointSet), Gaussian representation (G-Rep) is proposed to construct the Gaussian distribution using the MLE algorithm.
\item  To achieve an effective and robust measurement for the Gaussian distribution, three statistical distances, Kullback–Leibler divergence (KLD) \cite{kullback1951information}, Bhattacharyya Distance (BD) \cite{bhattacharyya1943measure}, and Wasserstein Distance (WD) \cite{villani2008optimal}, are explored and corresponding regression loss functions are designed and analyzed.
\item  To realize the consistency in the measurement between sample selection and loss regression, fixed and dynamic label assignment strategies are constructed based on a Gaussian metric to further boost performance.
\item  Extensive experiments were conducted on several publicly available datasets, e.g., DOTA, HRSC2016, UCAS-AOD, and ICDAR2015, and the results demonstrated the excellent performance of the proposed techniques for arbitrary-oriented object detection.
\end{itemize}

\section{Related Work} \label{section:Related Work}
\subsection{Oriented Object Representations}
The most popular representation of oriented object detection uses the five-parameter OBB, IoU value as the metric in label assignment, Smooth $l_{1}$ as the regression loss to regress the five parameters $(x, y, w, h, \theta)$ \cite{ ding2018learning, yang2021r3det, yang2018automatic, yang2020arbitrary,qian2021learning, yang2019scrdet}, respectively, where $(x,y)$, $w$, $h$, and $\theta$ denote the center coordinates, width, height, and angle of the box, respectively.
Additionally, some researchers have proposed methods using an eight-parameter QBB to represent the object and Smooth $l_{1}$ as regression loss to regress the four corner points of QBB \cite{xu2020gliding, ming2021optimization}.
The recent anchor-free method CFA \cite{guo2021beyond} uses flexible PointSet (i.e., point set) to represent the oriented objects, inspired by the horizontal object detection method RepPoints \cite{yang2019reppoints}.
In addition, other complex representations exist, such as polar coordinates \cite{zhou2020objects, zhao2021polardet} and middle lines \cite{wei2020oriented}.

\subsection{Regression Loss in Arbitrary-Oriented Object Detection}
The mainstream regression loss for OBB and QBB is Smooth $l_{1}$, which encounters boundary discontinuity and square-like problems. 
To solve these problems, SCRDet \cite{yang2019scrdet} and RSDet \cite{qian2021learning} adopt the IoU-Smooth $l_{1}$ loss and modulated loss to smooth the the boundary loss jump. CSL \cite{yang2020arbitrary} and DCL \cite{yang2021dense} transform angular prediction from regression to classification.
RIDet \cite{ming2021optimization} uses the representation invariant loss to optimize bounding box regression. 
Using the IoU value as the regression loss \cite{yu2016unitbox} has become a research topic of great interest in oriented object detection, which can avoid partial problems caused by the regression of the angle parameter or point ordering. 
Since then, many variants of the IoU loss have been developed, such as GIoU \cite{rezatofighi2019generalized}, DIoU \cite{zheng2020distance} and PIoU \cite{chen2020piou}.
However, at the same time, the losses of IoU series suffer from infeasible optimization, slow convergence speed, and gradient vanish in the case of non-overlapping bounding boxes.
GWD \cite{yang2021rethinking} and KLD \cite{yang2021learning} construct the Gaussian distribution for OBB, which overcome the dilemma of object detection and demonstrate the possibility of a unified solution for various issues.
However, aforementioned two works are limited to OBB representation, and are not truly unified solutions in terms of representation.

\subsection{Label Assignment Strategies}
Label assignment plays a vital role in detection performance, and many fixed and dynamic label assignment strategies have been proposed.
Classic object detection methods, such as Faster RCNN \cite{ren2015faster} and RetinaNet \cite{lin2017focal}, adopt a fixed max IoU strategy, which requires predefined thresholds for positive and negative samples in advance.
To overcome the difficulty in setting hyper-parameters, ATSS \cite{ATSS} uses statistical characteristics to calculate dynamic IoU thresholds.
Furthermore, PAA \cite{PAA} adaptively separates anchors into positive and negative samples for a ground-truth bounding box in a probabilistic manner.
Additionally, other excellent dynamic label assignment strategies exist. For instance, 
DAL \cite{ming2021dynamic} dynamically assigns labels according to a defined matching degree, and FreeAnchor \cite{FreeAnchor} dynamically selects labels under the maximum likelihood principle. 
Nevertheless, these methods still rely on the IoU value as the main indicator to evaluate the quality of the sample.

In addition, some excellent works focus on feature learning to improve detection results.
For example, Zhou et al. \cite{zhou2022interactive} proposed IRC that utilizes effective localization information to achieve learning in an interactive way with regression and classification branches.
Wang et al. \cite{wang2022crabnet} proposed a fully task-specific feature learning method CrabNet, which could learn disentangled features for classification and localization tasks in one-stage object detection methods.
Wang et al. \cite{wang2022robust} proposed an effective knowledge distillation framework FMD, which allows a thin student network to learn rich information in feature maps of a wide teacher network.
Wu et al. \cite{wu2022edn} proposed the EDN network, which was designed to use high-level features efficiently in the field of salient object detection.
Li et al. \cite{li2022boodet} proposed the BooDet method based on aggregating predictions to achieve high performance in object detection by bootstrap of classification and bounding box regression.

\section{Proposed Approach}
The main contribution of this paper focuses on three aspects.
First, Gaussian distribution is constructed for PointSet and QBB representation, which overcomes the dilemma shown in previous studies \cite{yang2021rethinking, yang2021learning} that the Gaussian distribution can only be applied to OBB.
Second, new regression loss functions are designed and analyzed for supervising network learning based on the Gaussian distribution. 
Third, the measurement between label assignment and loss regression based on the Gaussian distribution are aligned, and the new fixed and dynamic label assignment strategies are designed accordingly.

\figref{fig:overall} shows an overview of the proposed method. In the training phase, the predicted bounding box and ground truth box are converted into the Gaussian distribution in a manner according to their respective representations. 
For example, the MLE algorithm is used for the boxes in PointSet or QBB representations according to Equation~\eqref{equ:mle}, and the OBB uses the matrix conversion expressed in Equation~\eqref{equ:OBB2Gaussian1} or is converted to QBB and then uses MLE algorithm. 
Then, the devised fixed or dynamic G-Rep label assignment strategy is used to select samples based on Gaussian distance metrics (KLD, BD, or WD).
Finally, regression loss functions based on the Gaussian distance metrics are designed to minimize the distance between two Gaussian distributions. 
In the testing phase, the output in OBB form is obtained from the trained model, which is constructed based on the unified Gaussian representation.

\subsection{Object Representation as the Gaussian Distribution}
\textbf{PointSet.} RepPoints \cite{yang2019reppoints} is an anchor-free method, which is also a baseline method used in this paper.
RepPoints baseline is constructed with a backbone network, an initial detection head and a refined detection head.
The backbone network utilizes ResNet and Feature Pyramid Network (FPN) \cite{he2016deep, lin2017feature} to extract multi-scale features, and two detection heads consist of non-shared classification and regression sub-networks.
The object is represented as a set of adaptive sample points (i.e., PointSet) and the regression framework adopts deformable convolution \cite{dai2017deformable} for point learning.
The object is represented as PointSet $R$, which is defined as:
\begin{equation}
R =  \left\{ \left(x_i,  y_i \right)\right\}^K_{i=1},
\end{equation}
where $\left(x_i,  y_i \right)$ is the coordinates of the $i$-th point and $K$ is the point number of a point set, which is set to 9 by default.
In the refined detection head, the model predicts offset $(\Delta x_i, \Delta y_i)$ for refinement, and the new refined predicted point set can be simply expressed as:
\begin{equation}
R' =  \left\{ \left(x_i + \Delta x_i,  y_i + \Delta y_i \right)\right\}^K_{i=1}.
\end{equation}

\textbf{QBB.} The baseline with QBB representation is constructed on the anchor-based method Cas-RetinaNet proposed in \cite{ming2021optimization}, which contains a backbone network and two detection heads. 
QBB is defined as the four corner points of the object ($Q =  \left\{ \left(x^q_i,  y^q_i \right)\right\}^4_{i=1}$). 
Note that the four corner points of QBB must be sorted in advance to match the corners of the given ground truth representation one-to-one for regression in the original QBB baseline.
Additionally, from the definitions of the three representations, we can deduce that QBB can be regarded as a special case of PointSet, and OBB can be regarded as a special case of QBB.
Therefore, constructing the Gaussian distribution for PointSet is extremely generalized, which is also the major focus in this paper.

\begin{figure}[tb]
\centering
\includegraphics[width=0.9\linewidth]{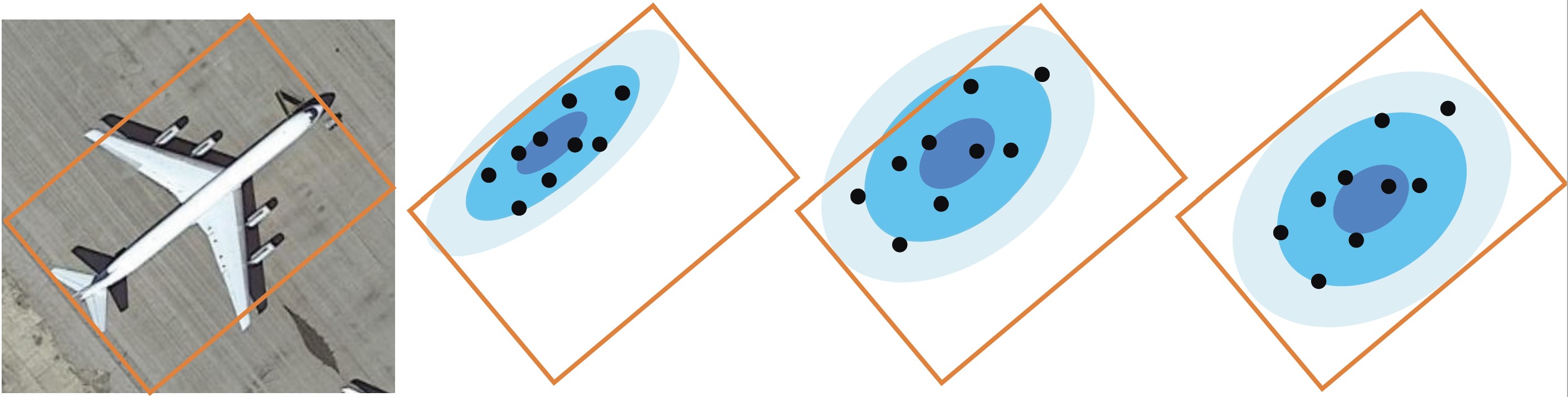}
\caption{Illustration of the learning process of the Gaussian distribution based on PointSet.} \label{fig:mle}
\end{figure}

\textbf{Transformation between PointSet/QBB and G-Rep.} Considering $(x_i,  y_i)$ as a two-dimensional (2-D) variable $\boldsymbol{x}_{i}$, its probability density under the Gaussian distribution $\mathcal{N}(\boldsymbol{\mu},\boldsymbol{\Sigma})$ is defined as
\begin{equation}
\mathcal{N}\left(\boldsymbol{x}_{i} \mid \boldsymbol{\mu}, \boldsymbol{\Sigma}\right) =\frac{\exp \left(-\frac{1}{2}\left(\boldsymbol{x}_{i}-\boldsymbol{\mu}\right)^\text{T} \boldsymbol{\Sigma}^{-1}\left(\boldsymbol{x}_{i}-\boldsymbol{\mu}\right)\right)}{2\pi \sqrt{\operatorname{det}\left(\boldsymbol{\Sigma}\right)}}.
\end{equation}
The parameters mean $\boldsymbol{\mu}$ and covariance matrix $\boldsymbol{\Sigma}$ of the Gaussian distribution are calculated using the maximum likelihood estimation (MLE) \cite{richards1961method} algorithm, which is expressed as:
\begin{equation}
\begin{aligned}
(\hat{\boldsymbol{\mu}}, \hat{\boldsymbol{\Sigma}})_{\mathrm{MLE}} 
&=\arg \max \prod_{i=1}^{N} \mathcal{N} \left(\boldsymbol{x}_{i}  \mid \boldsymbol{\mu}, \boldsymbol{\Sigma} \right)\\
&=\arg \max \sum_{i=1}^{N} \log \mathcal{N} \left(\boldsymbol{x}_{i}  \mid \boldsymbol{\mu}, \boldsymbol{\Sigma} \right),
\end{aligned}
\label{equ:mle}
\end{equation}

\begin{equation}
\hat{\boldsymbol{\mu}}=\frac{1}{N}\sum_{i=1}^{N}\boldsymbol{x}_{i}  \quad
\hat{\boldsymbol{\Sigma}}=\frac{1}{N}\sum_{i=1}^{N}(\boldsymbol{x}_{i}-\hat{\boldsymbol{\mu}})(\boldsymbol{x}_{i}-\hat{\boldsymbol{\mu}})^T,
\label{equ:mle_1}
\end{equation}
where $\boldsymbol{x}_{i}=\left(x_i, y_i \right)$ denotes the coordinates of the $i$-th point, and $N$ denotes the number of all points in a 2-D vector.
\figref{fig:mle} illustrates the Gaussian learning process based on PointSet. The MLE algorithm evaluates the parameters of the Gaussian distribution for the initialized PointSet, which is considered as a 2-D vector.
The coordinates of the points are updated according to the gradient feedback of the loss, and the Gaussian parameters are updated correspondingly.

\textbf{Transformation between OBB and G-Rep.}
In previous studies \cite{yang2021rethinking, yang2021learning}, a 2-D Gaussian distribution for OBB is constructed using a matrix transformation. 
For an OBB $(x,y,w,h,\theta)$, the parameters of the Gaussian distribution $\mathcal{N}\left(\boldsymbol{x} \mid \boldsymbol{\mu}, \boldsymbol{\Sigma}\right)$ are calculated as
\begin{equation}
\begin{aligned}
\boldsymbol{\mu}&=\left[x, y\right]^{\top}\\
\mathbf{\Sigma} &=\mathbf{R}_{\theta}
\boldsymbol{\Lambda} \mathbf{R}_{\theta}^{\top} \\
&=\left[\begin{array}{cc}
\cos \theta & -\sin \theta \\
\sin \theta & \cos \theta
\end{array}\right]\left[\begin{array}{cc}
\lambda_{1}& 0 \\
0 & \lambda_{2}
\end{array}\right]\left[\begin{array}{ccc}
\cos \theta & \sin \theta \\
-\sin \theta & \cos \theta
\end{array}\right] \\
&=\left[\begin{array}{cc}
\lambda_{1} \cos ^{2} \theta+\lambda_{2} \sin ^{2} \theta & (\lambda_{1}-\lambda_{2}) \cos \theta \sin \theta \\
(\lambda_{1}-\lambda_{2}) \cos \theta \sin \theta & \lambda_{1} \sin ^{2} \theta+\lambda_{2} \cos ^{2} \theta
\end{array}\right], \\
\end{aligned}
\label{equ:OBB2Gaussian1}
\end{equation}
where $\mathbf{R}_{\theta}$ and $\boldsymbol{\Lambda}$ are the rotation matrix and diagonal matrix of the eigenvalues, respectively.
The eigenvalues are calculated as  $\lambda_{1}=\frac{w^2}{4}$ and $\lambda_{2}=\frac{h^2}{4}$.

In summary, there are two Gaussian transformation methods adopted in this paper: MLE and matrix transformation.
The former can be used for conversion of all representations (OBB/QBB/PointSet), but is inefficient and inaccurate. The latter is more precise but only supports OBB. Therefore, when transforming ground truth, matrix transformation is chosen to avoid unnecessary bias.

The transformation from a Gaussian distribution to an OBB is necessary for calculating the mean average precision (mAP) in the testing process.
As shown in \figref{fig:overall}, prediction (QBB or PointSet) is obtained in the testing process.
Because network learning is supervised in the basis of the object Gaussian distribution, the prediction of the network output is also distributed in the form of the object Gaussian distribution. 
Therefore, it is necessary to convert the prediction (PointSet or QBB) into the Gaussian distribution, whose parameters are calculated using the MLE algorithm in Equation~\eqref{equ:mle}.
To go further, the OBB of the predicted object is obtained from the Gaussian distribution. 
For a predicted Gaussian distribution with known parameters $\boldsymbol{\mu}$ and $\boldsymbol{\Sigma}$, and the parameters of the corresponding OBB $(x, y, w, h, \theta)$ can also be obtained using the singular value decomposition (SVD) according to Equation~\eqref{equ:OBB2Gaussian1}.

\subsection{Gaussian Distance Metrics}\label{sec:GDM}
The keystone of an effective regression loss and label assignment is how to compute the similarity between the predicted Gaussian distribution $ \mathcal{N}_{p}\left( \boldsymbol{x}_{p} \mid \boldsymbol{\mu}_{p}, \boldsymbol{\Sigma}_{p}\right)$ and the ground-truth Gaussian distribution $ \mathcal{N}_{g}\left(\boldsymbol{x}_{g} \mid \boldsymbol{\mu}_{g}, \boldsymbol{\Sigma}_{g}\right)$.
Next, we will focus on three metrics to calculate the distance between two Gaussian distributions and further analyze their characteristics.

\textbf{Kullback–Leibler Divergence (KLD) \cite{kullback1951information}.}
The KLD between two Gaussian distributions is defined as
\begin{equation}
\begin{split}
 D_{K}\left(\mathcal{N}_{g}, \mathcal{N}_{p}\right) = & \frac{1}{2}\Big(\operatorname{tr}\left(\boldsymbol{\Sigma}_{p}^{-1} \boldsymbol{\Sigma}_{g} \right)  +  \ln \frac{\left|\boldsymbol{\Sigma}_{p}\right|}{\left|\boldsymbol{\Sigma}_{g}\right|}-2 \\
 & +\left(\boldsymbol{\mu}_{p}-\boldsymbol{\mu}_{g}\right)^{\top} \boldsymbol{\Sigma}_{p}^{-1}\left(\boldsymbol{\mu}_{p}-\boldsymbol{\mu}_{g}\right)\Big).  
\end{split}
\label{equ:KLD}
\end{equation}
Although KLD is not strictly a mathematical distance because it is asymmetric, KLD is scale invariant.
According to Equation~\eqref{equ:KLD} each item of KLD contains shape parameters $\boldsymbol{\Sigma}$ and center parameters $\boldsymbol{\mu}$.
All parameters form a chain coupling relationship and influence each other, which is very conducive to high-precision detection, as proved in \cite{yang2021learning}.

\textbf{Bhattacharyya distance \cite{bhattacharyya1943measure}.}
The BD between two Gaussian distributions is defined as
\begin{equation}
\begin{split}
 D_{B}\left(\mathcal{N}_{g}, \mathcal{N}_{p}\right)  = & \frac{1}{8}\left(\boldsymbol{\mu}_{g}-\boldsymbol{\mu}_{p}\right)^{T} \boldsymbol{\Sigma}^{-1}\left(\boldsymbol{\mu}_{g}-\boldsymbol{\mu}_{p}\right) \\
& +\frac{1}{2} \ln \bigg(\frac{\left| \boldsymbol{\Sigma}\right|}{\sqrt{\left|\boldsymbol{\Sigma}_{g} \boldsymbol{\Sigma}_{p}\right|}}\bigg),
\end{split}
\label{equ:BD}
\end{equation}
where $\quad \boldsymbol{\Sigma}=\frac{1}{2}\left(\boldsymbol{\Sigma}_{p}+\boldsymbol{\Sigma}_{g}\right)$.
Although BD is symmetric, it is not an actual distance either, because it does not satisfy the triangle inequality.
Similar with KLD, BD is also scale invariant.

\textbf{Wasserstein Distance (WD) \cite{villani2008optimal}.}
The WD between two Gaussian distributions is defined as
\begin{equation}
\begin{split}
D_{W} \left(\mathcal{N}_{g}, \mathcal{N}_{p}\right) = &
\left\|\boldsymbol{\mu}_{p}-\boldsymbol{\mu}_{q}\right\|^{2}+\operatorname{Tr}\left(\boldsymbol{\Sigma}_{p}\right)+\operatorname{Tr}\left(\boldsymbol{\Sigma}_{q}\right) \\
& -2 \operatorname{Tr}\left(\left(\boldsymbol{\Sigma}_{p}^{1 / 2} \boldsymbol{\Sigma}_{q} \boldsymbol{\Sigma}_{p}^{1 / 2}\right)^{1 / 2}\right).
\end{split}
\label{equ:WD}
\end{equation}
Different from KLD and BD, WD is an actual mathematical distance, which satisfies the triangle inequality and is symmetric. Note that WD is mainly divided into two parts: the distance between the center points ($x, y$) and and the coupling terms about $h$, $w$, and $\theta$. Although WD can greatly improve the performance of high-precision rotation detection because of the coupling between parts of the parameters, the independent optimization of the center point slightly shifts the detection result \cite{yang2021learning}.

\subsection{Regression Loss Based on Gaussian Metric}
The actual value obtained by the distance metric between Gaussian distributions is too large to be the regression loss, which leads to convergence difficulties.
Therefore, normalization is necessary so that the Gaussian distance can be used as regression loss.
The normalized functions are designed by a series of small-scale experiments, in which some empirical functions are tested, and the best functions for different metrics are chosen according to the results. 
Details are described in \secref{sec:NormalizedFunctionDesign}.
The normalized regression loss functions for KLD, BD, and WD are defined as
\begin{equation}
\mathcal{L}_\text{KLD} =1-\frac{1}{2 + \sqrt{D_{K}}\left(\mathcal{N}_{g}, \mathcal{N}_{p}\right)},
\end{equation}
\begin{equation}
\mathcal{L}_\text{BD} =1- \frac{1}{1 +  D_{B}\left(\mathcal{N}_{g}, \mathcal{N}_{p}\right)},
\end{equation}
\begin{equation}
\mathcal{L}_\text{WD} =1-\frac{1}{1 + \log(1 + D_{W} \left(\mathcal{N}_{g}, \mathcal{N}_{p}\right))},
\end{equation}
where $D_{K}\left(\mathcal{N}_{g}, \mathcal{N}_{p}\right)$, $D_{B}\left(\mathcal{N}_{g}, \mathcal{N}_{p}\right)$, and $D_{W} \left(\mathcal{N}_{g}, \mathcal{N}_{p}\right)$ are defined in Equation~\eqref{equ:KLD}, Equation~\eqref{equ:BD} and  Equation~\eqref{equ:WD}, respectively.

Recall that the most common regression loss for object detection is the Smooth $l_{1}$ loss.
Smooth $l_{1}$ loss independently minimizes each parameter of the prediction.
Taking the OBB as an example, the center point $(x, y)$, the weight $w$, the height $h$, and the angle $\theta$ are optimized separately. 
In fact, different objects have different sensitivities to different parameters. For instance, objects with large aspect ratios are sensitive to the angle error.
Hence, decoupled regression way is difficult to handle various objects with diverse shapes.  
For different representations, the Smooth $l_{1}$ loss suffers from different issues, as illustrated in \figref{fig:questions}. 
By contrast, all three Gaussian metric-based regression losses avoid issues of the boundary discontinuity, square-like problem, representation ambiguity, and isolated points.
Among them, WD is strictly the actual distance because of its triangle equality and symmetry, whereas KLD is asymmetric and BD satisfies the triangle inequality.
Although the three proposed regression loss functions have different characteristics, the overall difference in detection performance is slight and all are superior to the traditional Smooth $l_{1}$ or IoU loss.

\subsection{Label Assignment Based on Gaussian Metric}
Label assignment strategy is another key task for object detection. 
The most popular label assignment strategy is the IoU-based strategy, which assigns a label by comparing IoU values (proposals and ground truth) with IoU threshold.  
However, if the regression loss is based on a Gaussian distribution, inconsistencies can occur when the metrics of label assignment and regression loss are different. Therefore, new label assignment strategies based on the Gaussian distribution are devised. 
To go further, dynamic label assignment strategies with better adaptability and fewer hyper-parameters are also designed for G-Rep.

\textbf{Fixed G-Rep Label Assignment.} 
The range of the IoU value is $\left[0, 1\right]$ according to the definition of IoU, and the threshold values are selected empirically in the range $\left[0.3, 0.7\right]$. 
However, this strategy is clearly not applicable to the Gaussian distribution distance calculated by the three metrics described in \secref{sec:GDM}, whose value ranges are not closed intervals.
Along with the concept of G-Rep regression loss design, normalized functions for each distance evaluation metric are adopted. 
Additionally, the key formula for normalization function design maps the evaluation values to $\left[0, 1\right]$. In this paper, functions are explored for normalizing the obtained distance metrics (KLD, BD, and WD) in the label assignment process, which are listed as follows:
\begin{equation}
\mathcal{S}_{\text{KLD}}=\frac{1}{2+D_{K}\left(\mathcal{N}_{g}, \mathcal{N}_{p}\right)},
\label{equ:KLDinLA}
\end{equation}
 \begin{equation}
\mathcal{S}_{\text{BD}}=\frac{1}{1+D_{B}^{2}\left(\mathcal{N}_{g}, \mathcal{N}_{p}\right)},
\label{equ:BDinLA}
\end{equation}
 \begin{equation}
\mathcal{S}_{\text{WD}}=\frac{1}{2+D_{W} \left(\mathcal{N}_{g}, \mathcal{N}_{p}\right)},
\label{equ:WDinLA}
\end{equation} 
where $\mathcal{S}$ denotes the normalized metric for evaluating the similarity between the Gaussian distribution of a proposal and the ground truth. 
The optimal hyper-parameters require a empirical and experimental optimization. 
A series of experiments are conducted with unified positive threshold of 0.4 and negative threshold of 0.3, respectively. The results are listed in \tabref{tab:LossHRSCRes}, where rows 2–4 demonstrate that optimal hyper-parameter thresholds are difficult to set for different distance metrics; hence, dynamic label assignment strategies based on a Gaussian are further explored.

\textbf{Dynamic G-Rep Label Assignment.}
Dynamic G-Rep label assignment strategies are devised based on the three distance metrics in \secref{sec:GDM} to avoid the difficulty of selecting the optimal hyper-parameters.
Inspired by ATSS \cite{ATSS}, the threshold for selecting positive and negative samples is calculated dynamically according to the statistical characteristics of all the normalized distances (calculated in Equation~\eqref{equ:KLDinLA}, Equation~\eqref{equ:BDinLA}, and Equation~\eqref{equ:WDinLA}).
For the $i$-th ground truth, the dynamic threshold $\mathcal{T}$ is calculated as
\begin{equation}
\begin{aligned}
\mathcal{T}_{i} = \frac{1}{J}  \sum_{j=1}^{J}{\mathcal{I}_{i,j}} 
 + \sqrt{\frac{1}{J}  \sum_{j=1}^{J}{(\mathcal{I}_{i,j}
 - \frac{1}{J} \sum_{j=1}^{J}{\mathcal{I}_{i,j}})^{2}}}, 
\end{aligned}
\end{equation}
where $J$ is the number of candidate samples, and $\mathcal{I}_{i,j}$ is the normalized value of KLD, BD, or WD between the $i$-th ground truth and the $j$-th proposal, which is calculated in Equation~\eqref{equ:KLDinLA}, Equation~\eqref{equ:BDinLA}, or Equation~\eqref{equ:WDinLA}.
Next, positive samples are selected using the general assignment strategy, that is, candidates are selected whose similarity values are greater than or equal to the threshold  $\mathcal{T}_{i}$.

Furthermore, motivated by PAA \cite{PAA}, another approach using Gaussian distributions in the label assignment process is adopted in this paper.
Specifically, each sample is assigned a score for quality evaluation.
The Gaussian mixture model (GMM) with two components is adopted to model the score distribution of samples, and the parameters of the GMM are calculated using the expectation-maximization algorithm \cite{1977Maximum}. 
Finally, samples are assigned labels according to their probabilities. PAA and ATSS are modified and combined to construct the robust dynamic label assignment strategy PATSS in the proposed method.

\section{Experiments} 
\subsection{Datasets and Implementation Details}
Experiments are conducted on the public aerial image datasets DOTA \cite{xia2018dota}, HRSC2016 \cite{liu2017high}, UCAS-AOD \cite{zhu2015orientation}, and scene text dataset ICDAR2015 \cite{karatzas2015icdar} to verify the superiority of the proposed method.

DOTA \cite{xia2018dota} is a public oriented object detection benchmark dataset for aerial images, which contains 15 object categories: plane (PL), baseball diamond (BD), bridge (BR), ground track field (GTF), small vehicle (SV), large vehicle (LV), ship (SH), tennis court (TC), basketball court (BC), storage tank (ST),  soccer ball field (SBF), roundabout (RA), harbor (HA), swimming pool (SP), and helicopter (HC).
In the experiments, DOTA’s official training and validation dataset sets are selected as the training set, whose images are split into blocks of 1024$\times$1024 pixels with an overlap of 200 pixels and scaled to 1333$\times$1024 for training.

HRSC2016 \cite{liu2017high} is a public aerial image dataset for ship detection.
The number of images in the training, variation, and testing sets are 436, 181 and 444, respectively.

The public aerial image dataset UCAS-AOD \cite{zhu2015orientation} is another multi-class oriented object detection dataset, and contains two categories: car and airplane. 

ICDAR2015 \cite{karatzas2015icdar} is commonly used for oriented scene text detection and spotting. This dataset contains 1000 training images and 500 testing images. 

Experiments are conducted using the MMDetection framework \cite{chen2019mmdetection} (under the Apache-2.0 License) on 2 Titan V GPUs with 11GB memory.
Two baselines of RepPoints with PointSet and Cas-RetinaNet with QBB are conducted.

The optimizers of the RepPoints and Cas-RetinaNet frameworks are SGD with a 0.01 initial learning rate and Adam with a 0.0001 initial learning rate, respectively.

The framework for the RepPoints baseline is trained for 12 and 36 epochs on DOTA and HRSC2016 for the ablation experiments, and 40, 80, 120 and 240 epochs on DOTA, HRSC2016, UCAS-AOD and ICDAR2015 with data augmentation, respectively. 
The framework for the Cas-RetinaNet baseline is trained for 12, 100, 36 and 100 epochs on DOTA,  HRSC2016, UCAS-AOD and ICDAR2015, respectively.

The mAP is adopted as evaluation metric for testing results on DOTA, HRSC2016 and UCAS-AOD. Precision, recall, and F-measure (i.e., F1) are adopted on ICDAR2015 following official criteria.

\subsection{Normalized Function Design} \label{sec:NormalizedFunctionDesign}

\begin{table}[htb]
\caption{Experiment results of normalized function design for label assignment ($\mathcal{S}$) and regression loss ($\mathcal{L}$) on HRSC2016.} \label{tab:FunHRSCRes} 
\begin{center}\footnotesize
   \begin{tabular}{c|c|c|c}
    \hline
    \textbf{Metric} & \textbf{Fun. of $\mathcal{S}$ } & \textbf{Fun. of $\mathcal{L}$} & \textbf{mAP(\%)} \\
    \hline
    \multirow{6}{*}{KLD}
    & $\frac{1}{2+D_{K}}$ & $1- \frac{1}{2 + \sqrt{D_{K}}}$ & \textbf{88.06} \\  \cline{2-4}
    & $\frac{1}{2+D_{K}}$ & $1- e^{-\sqrt{D_{K}}}$ & 87.32 \\ \cline{2-4}
    & $\frac{1}{2+D_{K}}$ & $1-e^{-{D^2_{K}}}$ & 50.73  \\ \cline{2-4}
    & $\frac{1}{2+\sqrt{D_{K}}}$ & $1- \frac{1}{2 + \sqrt{D_{K}}}$ & 35.46 \\ \hline
    \multirow{6}{*}{BD}
    & $\frac{1}{1+D^2_{B}}$ & $1- \frac{1}{1+ D_{B}}$ & \textbf{85.32} \\ \cline{2-4}
    & $\frac{1}{1+D^2_{B}}$ &$ \log(1+D_{B}$) & 85.02 \\ \cline{2-4}
    & $\frac{1}{1+D^2_{B}}$ & $5\times D_{B}$ & 60.68 \\ \cline{2-4}
    & $\frac{1}{1+D_{B}}$ & $1-\frac{1}{1+ D_{B}}$ & 85.12 \\ \hline
    \multirow{6}{*}{WD}
    & $\frac{1}{2+D_{W}}$ & $1-\frac{1}{1+ \log(1+D_{W})}$ & \textbf{88.56} \\ \cline{2-4}
    & $\frac{1}{2+D_{W}}$ & $1- \frac{1}{2 + \sqrt{D_{W}}}$ & 87.04 \\ \cline{2-4}
    & $\frac{1}{2+D_{W}}$ & $1- e^ {-\sqrt{D_{W}}}$ & 88.24 \\ \cline{2-4}
    & $\frac{1}{2+\sqrt{D_{W}}}$ &
    $1-\frac{1}{1+\log(1 + D_{W})}$
    & 87.54  \\ \hline
    \end{tabular}
\end{center}
\end{table}

\tabref{tab:FunHRSCRes} lists the experimental results of the normalized function design. The design principle of the normalized functions for label assignment is mapping the calculated values of Gaussian statistical distance to the values in $[0,1]$. The design principle of the normalized functions for regression loss is transforming the calculated values to the values in a suitable range (e.g., $[0, 10]$).
In addition, the normalized regression loss functions should have a reasonable gradient descent.
Some generic functions, such as $\log(*)$ and $e^*$, are selected in the experiments. 
Finally, the appropriate normalized functions are selected according to the best results of the experiments.

\subsection{Ablation Study}

\begin{table}[htb]
\caption{Ablation study of G-Rep for PointSet on DOTA and HRSC2016. $\mathcal{S}$ and $\mathcal{L}$ represent the label assignment strategy and regression loss function, respectively.} \label{tab:PointSetDotaRes}
\begin{center}\footnotesize
    \begin{tabular}{c|cccc}
    \hline
     \textbf{Data} & \textbf{Representation} & \textbf{$\mathcal{S}$ } & \textbf{$\mathcal{L}$} & \textbf{Evaluation} \\ \hline
     \multirow{6}{*}{DOTA} &
     PointSet & IoU (Max)  & GIoU & 63.97 \\
    \cline{2-5}
     &  \multirow{2}{*}{G-Rep} 
     & IoU (Max) &  $\mathcal{L}_\text{KLD}$ & 64.63 \textbf{(+0.66)}\\
     & & $\mathcal{S}_{\text{KLD}}$ (Max)   & $\mathcal{L}_\text{KLD}$ & 65.07 \textbf{(+1.10)} \\
     
     \cline{2-5}
     & PointSet & IoU (ATSS)  & GIoU & 68.88 \\
      \cline{2-5}
     &\multirow{2}{*}{G-Rep}  & $\mathcal{S}_{\text{KLD}}$ (ATSS)  & $\mathcal{L}_{\text{KLD}}$ & 70.45 \textbf{(+1.57)} \\
     & & $\mathcal{S}_{\text{KLD}}$ (PATSS) & $\mathcal{L}_\text{KLD}$ & 72.08 \textbf{(+3.20)}\\
    \hline
    \multirow{3}{*}{HRSC2016} 
     & PointSet & IoU (ATSS)  & GIoU & 78.07 \\
     \cline{2-5}
    &\multirow{2}{*}{G-Rep}  
    & $\mathcal{S}_{\text{KLD}}$ (ATSS)  & $\mathcal{L}_{\text{KLD}}$ & 88.06 \textbf{(+9.99)} \\
     & & $\mathcal{S}_{\text{KLD}}$ (PATSS) & $\mathcal{L}_\text{KLD}$ & 89.15 \textbf{(+11.07)}\\
     \hline
    \end{tabular}
\end{center}
\end{table}

\tabref{tab:PointSetDotaRes} compares the performance of the G-Rep and PointSet on the DOTA dataset.
``IoU (Max)'' and ``IoU (ATSS)'' represent the fixed predefined threshold strategy and dynamic ATSS \cite{ATSS} label assignment strategy based on the IoU metric, respectively.
The baseline method of PointSet is RepPoints \cite{yang2019reppoints} with ResNet50-FPN \cite{he2016deep, lin2017feature}. 
Note that the IoU for the Gaussian distribution is calculated as the IoU between the box transformed by the Gaussian distribution and ground truth.
The predefined positive and negative thresholds in the fixed strategy for IoU and $\mathcal{S}_{\text{KLD}}$ are 0.4 and 0.3, respectively.
The superiority of G-Rep is reflected in two aspects: regression loss and label assignment.

\textbf{Analysis of regression loss based on G-Rep.} Even if only the GIoU is replaced by $\mathcal{L}_\text{KLD}$, the performance of G-Rep is better than that of PointSet (\textbf{64.63\%} vs. \textbf{63.97\%}).
The dynamic label assignment strategies avoid the influence of unsuitable hyper-parameters for a fair comparison of the GIoU and Gaussian regression loss.
Additionally, the superiority of G-Rep is clearly demonstrated when the dynamic label assignment strategies are used. 
$\mathcal{L}_\text{KLD}$ still surpassed the GIoU with the same dynamic label assignment strategy ATSS on DOTA (\textbf{70.45\%} vs. \textbf{68.88\%}).

\textbf{Analysis of label assignment based on G-Rep.} 
The label assignment strategy is another important factor for high-detection performance. 
For the $\mathcal{L}_\text{KLD}$ loss, \tabref{tab:PointSetDotaRes} shows the detection results of different label assignment strategies.
Using KLD performs better than using IoU as a metric of the label assignment, which demonstrates the effectiveness of aligning the label assignment and regression loss metrics.
The optimal fixed negative and positive thresholds for selecting samples are difficult to select, whereas dynamic label assignment strategies avoid this issue.
PATSS denotes the combination of the ATSS \cite{ATSS} and PAA \cite{PAA} strategies.
The mAP further reached \textbf{70.45\%} and \textbf{72.08\%} under the more robust dynamic selection strategies ATSS and PATSS, respectively.
Without additional features, the combination of the dynamic label assignment strategy and regression loss increased the mAP by \textbf{8.11\%} compared with the baseline method.

\textbf{Analysis of the advantages for an object with a large aspect ratio.}
Outlier results in more serious location errors for objects with large aspect ratios than square objects.
\tabref{tab:LarRes} shows that G-Rep is more effective than PointSet for the objects with large aspect ratio, where the mAP increases by \textbf{6.18\%} for the five typical categories with narrow objects on DOTA, because G-Rep is not sensitive to isolated points.

\begin{table}[htb]
\caption{Performance comparison of PointSet and G-Rep for large aspect ratio objects. `mAR' represents the mean aspect ratio (the ratio of long side to short side) of all targets in one category. }\label{tab:LarRes}
\begin{center}
\resizebox{\textwidth}{!}{
\centering
\begin{tabular}{c|ccccc|c}
\hline
\multirow{2}{*}{\textbf{Rep.}}  & \textbf{BR} & \textbf{SV} & \textbf{LV} & \textbf{SH} & \textbf{HC} & \multirow{2}{*}{\textbf{\textbf{mAP(\%)}}} \\ 
 \cline{2-6}  & mAR=2.93 & mAR=1.72 & mAR=3.45 & mAR=2.40 & mAR=2.34 & \\
\hline
PointSet & 46.87 & 77.10 & 71.65 & 83.71 & 32.93  & 62.45 \\
G-Rep & 50.82 \textbf{(+3.95)} & 79.33  \textbf{(+2.23)} & 75.07 \textbf{(+3.51)} & 87.32 \textbf{(+3.61)} & 50.63 \textbf{(+17.70)} & 68.63 \textbf{(+6.18)} \\
\hline
\end{tabular}}
\end{center}
\end{table}

\begin{table}[htb]
\caption{Comparison of the three Gaussian distances as metrics for label assignment and regression loss on HRSC2016.} \label{tab:LossHRSCRes}
\begin{center}\footnotesize
\begin{tabular}{c|ccc}
\hline
\textbf{Representation}  & \textbf{$\mathcal{S}$}  & \textbf{$\mathcal{L}$} & \textbf{mAP(\%)} \\ \hline
 PointSet & IoU (ATSS) & GIoU & 78.07 \\
\hline 
\multirow{10}{*}{G-Rep} 
 & $\mathcal{S}_{\text{KLD}}$ (Max)  & $\mathcal{L}_{\text{KLD}}$ & 73.44 \\
&  $\mathcal{S}_{\text{BD}}$ (Max)   & $\mathcal{L}_{\text{BD}}$ &  46.71 \\
&  $\mathcal{S}_{\text{WD}}$ (Max)   & $\mathcal{L}_{\text{WD}}$ & 84.39 \\ 
 \cline{2-4}
& $\mathcal{S}_{\text{KLD}}$ (ATSS)  & $\mathcal{L}_{\text{KLD}}$ & 88.06 \\
&  $\mathcal{S}_{\text{BD}}$ (ATSS)   & $\mathcal{L}_{\text{BD}}$ & 85.32 \\
&  $\mathcal{S}_{\text{WD}}$ (ATSS)   & $\mathcal{L}_{\text{WD}}$ & 88.56 \\  
\cline{2-4}
&  $\mathcal{S}_{\text{KLD}}$ (ATSS)    & $\mathcal{L}_{\text{BD}}$ & 88.90 \\
& $\mathcal{S}_{\text{KLD}}$ (ATSS)    & $\mathcal{L}_{\text{WD}}$ &  88.80 \\
&  $\mathcal{S}_{\text{BD}}$ (ATSS)  & $\mathcal{L}_{\text{KLD}}$ &  85.32 \\
&  $\mathcal{S}_{\text{BD}}$ (ATSS)  & $\mathcal{L}_{\text{WD}}$ &  85.28 \\
\hline
\end{tabular}
\end{center}
\end{table}

\textbf{Comparison of different Gaussian distance metrics.}
\tabref{tab:LossHRSCRes} compares the performances when different evolution metrics, KLD, WD and BD, are used in fixed and dynamic label assignment strategies and regression loss.
The performances based on fixed label assignment strategies varied greatly as a result of the hand-crafted hyper-parameters. 
Therefore, experiments based on dynamic label assignment strategies are constructed to objectively compare the performances of the metrics. 
The experimental results demonstrate that the overall performance of the G-Rep loss functions surpasses that of the GIoU loss. 
There are tolerable performance differences between BD and the other two losses, and a slight difference (within 0.5\%) between KLD and WD. 
To further explore whether KLD and WD are more suitable as the regression loss than BD, the label assignment metrics are unified as KLD (rows 5, 8 and 9) for the ablation study of the loss functions. 
In fact, all the three G-Rep losses outperform the baseline (RepPoints) \cite{yang2019reppoints}. 
There are slight differences between them in detection performance.

\begin{table}[htb]
\caption{Ablation study of G-Rep for QBB representations on various datasets. `*' denotes that dynamic ATSS-based strategies are adopted. The regression loss of G-Rep is the $\mathcal{L}_{\text{KLD}}$.} \label{tab:RepresentationsHRSCRes}
\begin{center}\footnotesize
\begin{tabular}{c|ccc}
\hline
  \textbf{Data} &  \textbf{Representation} & \textbf{Eval.} &  \textbf{Gain $\uparrow$} \\ 
\hline
\multirow{4}{*}{DOTA}  
& PointSet* & 68.88 & -- \\
 &G-Rep* (PointSet) & 70.45&  \textbf{+1.57} \\
& QBB &63.05& --\\
& G-Rep (QBB)   &  67.92 & \textbf{+4.87} \\
  \cline{2-4}
\hline

\multirow{4}{*}{HRSC2016}  
& PointSet* & 78.07 & -- \\
 &G-Rep* (PointSet) & 88.06 & \textbf{+9.99}\\
& QBB &87.70 & --\\
& G-Rep (QBB)   &88.01 & \textbf{+0.31} \\
 \cline{2-4}

\hline

\multirow{4}{*}{UCAS-AOD}  
& PointSet* & 90.15 & -- \\
 &G-Rep* (PointSet) & 90.20 & \textbf{+0.05}   \\
& QBB & 88.50 & --\\
& G-Rep (QBB)   &88.82 & \textbf{+0.32} \\
  \cline{2-4}

\hline
\multirow{4}{*}{ICDAR2015}  
& PointSet* & 76.20 & -- \\
 &G-Rep* (PointSet) & 81.30 & \textbf{+5.10} \\
& QBB &75.10& -- \\
& G-Rep (QBB)   & 75.83 & \textbf{+0.73} \\
  \cline{2-4}
\hline
\end{tabular}
\end{center}
\end{table}

\textbf{Ablation study on various datasets.}
\tabref{tab:RepresentationsHRSCRes} shows the experimental results of G-Rep using two baselines on various datasets. 
The QBB baseline adopted the anchor-based method Cas-RetinaNet \cite{ming2021optimization} (i.e., the cascaded RetinaNet \cite{lin2017focal}).
G-Rep resulted in varying degrees of improvement on the anchor-based baseline with QBB and anchor-free baseline with PointSet on various datasets.
There are three key aspects to these results:
\begin{enumerate}
    \item \textbf{Elongated objects}. On the datasets containing a large number of elongated objects (e.g., HRSC2016, ICDAR2015), the improvement of G-Rep applied to PointSet is more pronounced than that applied to QBB, as shown in \tabref{tab:LarRes}, mainly because that the greater the number of points, the more accurate the Gaussian distribution obtained, and thus the more accurate the representation of the elongated object. 
    \item \textbf{Size of dataset}. The performance on the small datasets (e.g., UCAS-AOD) tends to be saturated, so the improvement is relatively small.
    \item \textbf{High baseline}. Models with high performance baseline are hard to improve significantly (e.g., HRSC2016-QBB, DOTA-PointSet).
\end{enumerate}

\subsection{Time Cost Analysis}

\tabref{tab:time_analysis} compares the time cost of G-Rep and other two methods. 
Since their parameters are the same, the main difference during inference lies in the post-processing. Although MLE slow down the inference speed, G-Rep achieves a performance improvement of 5.17\% over baseline, which is worthwhile. In addition, G-Rep achieves better performance with fewer parameters based on RepPoints (anchor-free) compared to S$^2$ANet (anchor-based).

\begin{table}[htb]
  \caption{Comparison of the time cost on DOTA.}
  \label{tab:time_analysis}
  \begin{center}\footnotesize
  \begin{tabular}{c|c|c|c}
        \toprule
        Method & mAP &  Params & Speed \\
        \hline
        \multirow{1}{*}{RepPoints} & 70.39 & \textbf{36.1M} & \textbf{24.0fps} \\ 
        \multirow{1}{*}{S$^2$ANet} & 74.12 & 37.3M & 19.9fps \\ 
        \multirow{1}{*}{\textbf{G-Rep (ours)}} & \textbf{75.56} & \textbf{36.1M} & 19.3fps \\     
        \bottomrule
    \end{tabular}
    \end{center}
\end{table}

\begin{table}[H]
\caption{Comparison of various detectors of $\mathrm{mAP}$ values on the OBB-based task of the DOTA-v1.0. MS indicates that multi-scale training.} \label{tab:DOTARes}
\vskip 2ex
\hskip -0.1\textwidth
\resizebox{1.2\textwidth}{!}{
\tabcolsep=2.5pt
\centering
\begin{tabular}{c|c|c|c|c|c|c|c|c|c|c|c|c|c|c|c|c|c|c}
\hline
\textbf{Method}&\textbf{Backbone} & \textbf{MS} & \textbf{PL}& \textbf{BD}& \textbf{BR}& \textbf{GTF}& \textbf{SV}& \textbf{LV}& \textbf{SH}& \textbf{TC}& \textbf{BC}& \textbf{ST}& \textbf{SBF}& \textbf{RA}& \textbf{HA}& \textbf{SP}& \textbf{HC} &\textbf{mAP(\%)}\\ \hline
\emph{two-stage:} \\ 
\hline
ICN \cite{azimi2018towards}  &R-101 &$\checkmark$ & 81.40 & 74.30 & 47.70 & 70.30 & 64.90 & 67.80 & 70.00 & 90.80 & 79.10 & 78.20 & 53.60 & 62.90 & 67.00 & 64.20 & 50.20 & 68.20 \\ 
GSDet \cite{li2021GSDet}  &R-101& & 81.12 &76.78& 40.78& 75.89 &64.50& 58.37& 74.21 &89.92& 79.40& 78.83& 64.54 &63.67& 66.04 &58.01 &52.13& 68.28 \\
RADet \cite{li2020radet} &RX-101 & $\checkmark$& 79.45 &76.99 &48.05&65.83&65.45&74.40&68.86&89.70&78.14&74.97&49.92&64.63&66.14&71.58&62.16&69.06 \\ 
RoI-Transformer \cite{ding2018learning} &R-101&$\checkmark$  & 88.64 & 78.52 & 43.44 & 75.92 & 68.81 & 73.68 & 83.59 & 90.74 & 77.27 & 81.46 & 58.39 & 53.54 & 62.83 & 58.93 & 47.67 & 69.56 \\  
CAD-Net \cite{zhang2019cad} &R-101&   & 87.80 & 82.40 & 49.40 & 73.50 & 71.10 & 63.50 & 76.70 & \textbf{90.90} & 79.20 & 73.30 & 48.40 & 60.90 & 62.00 & 67.00 & 62.20 & 69.90\\  

SCRDet \cite{yang2019scrdet} &R-101&$\checkmark$ & 89.98&80.65&52.09&68.36&68.36&60.32&72.41&90.85&\textbf{87.94}&86.86&65.02& 66.68 &66.25&68.24&65.21&72.61\\  
SARD \cite{wang2019sard} &R-101& &89.93 &84.11&54.19&72.04 &68.41 &61.18 &66.00 &90.82 & 87.79 &86.59 &65.65 &64.04 & 66.68 &68.84 &68.03& 72.95 \\  
FADet \cite{li2019feature} &R-101& $\checkmark$&90.21 &79.58 &45.49 &76.41 & 73.18 &68.27 &79.56 &90.83 &83.40 &84.64 &53.40 &65.42 & 74.17 &69.69 &64.86 & 73.28 \\  
MFIAR-Net\cite{yang2020multi} &R-152& $\checkmark$& 89.62& 84.03& 52.41 & 70.30& 70.13&67.64 &77.81 &90.85 &85.40 &86.22&63.21 &64.14&68.31&70.21&62.11&73.49\\ 
Gliding Vertex \cite{xu2020gliding} &R-101& & 89.64& 85.00& 52.26& 77.34 & 73.01& 73.14& 86.82& 90.74& 79.02& 86.81& 59.55& \textbf{70.91}& 72.94& 70.86 & 57.32 & 75.02 \\ 
CenterMap \cite{wang2020learning} & R-101 &$\checkmark$ & 89.83 & 84.41 &  54.60 & 70.25 & 77.66& 78.32& 87.19& 90.66& 84.89& 85.27& 56.46& 69.23& 74.13& 71.56& 66.06 & 76.03 \\

CSL (FPN-based) \cite{yang2020arbitrary}& R-152 &$\checkmark$& \textbf{90.25}& 85.53& 54.64 &75.31& 70.44& 73.51& 77.62& 90.84& 86.15& 86.69& 69.60& 68.04& 73.83& 71.10& 68.93 &76.17 \\
RSDet \cite{qian2021learning} & R-152 & $\checkmark$ & 89.93 & 84.45 & 53.77& 74.35& 71.52& 78.31& 78.12& 91.14& 87.35& 86.93& 65.64& 65.17& 75.35&  79.74 & 63.31& 76.34 \\
OPLD \cite{song2020learning} & R-101& $\checkmark$ &89.37 &\textbf{85.82}& 54.10 &\textbf{79.58} &75.00 &75.13 &86.92& 90.88& 86.42& 86.62& 62.46& 68.41 &73.98& 68.11& 63.69&76.43 \\
SCRDet++ \cite{yang2020scrdet++}& R-101 & $\checkmark$ &90.05& 84.39& 55.44& 73.99& 77.54& 71.11& 86.05& 90.67& 87.32& 87.08&  69.62& 68.90& 73.74& 71.29 &65.08& 76.81 \\
\hline 
\emph{one-stage:}  \\
\hline
 $ \text{P-RSDet}$ \cite{zhou2020objects} & R-101 & & 89.02&73.65&47.33& 72.03 &70.58&73.71&72.76&90.82&80.12&81.32&59.45&57.87&60.79&65.21&52.59& 69.82 \\ 
$ \text{O}^{2}\text{-Det}$ \cite{wei2020oriented} & H-104 & &89.31 & 82.14 &47.33& 61.21 &71.32 & 74.03 &78.62&90.76 &82.23&81.36& 60.93&60.17&58.21&66.98&61.03&71.04 \\ 

ACE \cite{dai2022ace} & DAL34\cite{yu2018deep}  & & 89.50 &76.30& 45.10& 60.00& 77.80& 77.10 &86.50& 90.80& 79.50& 85.70& 47.00& 59.40& 65.70 &71.70 &63.90 &71.70 \\

$ \text{R}^{3}\text{Det}$ \cite{yang2021r3det} & R-152& $\checkmark$& 89.24 &80.81&51.11&65.62&70.67&76.03&78.32&90.83&84.89&84.42&65.10& 57.18&68.10&68.98&60.88&72.81 \\  
\text{BBAVectors} \cite{yi2021oriented} & R-101 & $\checkmark$ & 88.35 &79.96 & 50.69 & 62.18 &78.43 &78.98 &87.94 &90.85 &83.58 &84.35 &54.13& 60.24 &65.22& 64.28& 55.70 &73.32 \\
DRN \cite{pan2020dynamic}  & H-104 & $\checkmark$ & 89.71 &82.34& 47.22& 64.10& 76.22& 74.43& 85.84& 90.57 &86.18 &84.89& 57.65& 61.93& 69.30& 69.63& 58.48& 73.23 \\
GWD \cite{yang2021rethinking} & R-152 & & 88.88 &   80.47 & 52.94 & 63.85 & 76.95 & 70.28 & 83.56 & 88.54 & 83.51 & 84.94 & 61.24 & 65.13 & 65.45 & 71.69 & 73.90 & 74.09\\

RO$^{3}$D \cite{hou2022refined} & R-101 &$\surd$ &88.69 &79.41 & 52.26 &65.51 &74.72&80.83 &87.42 &90.77&84.31 &83.36 &62.64 &58.14&66.95&72.32 & 69.34 & 74.44 \\
 
CFA \cite{guo2021beyond} & R-101 &  & 89.26 & 81.72& 51.81& 67.17 & 79.99 & 78.25 & 84.46 & 90.77 &83.40&85.54&54.86&67.75&73.04&70.24&64.96&75.05 \\
KLD \cite{yang2021learning} & R-50 & & 88.91 & 83.71 & 50.10 & 68.75 & 78.20 & 76.05 & 84.58 & 89.41 & 86.15 & 85.28 & 63.15 & 60.90 &75.06 & 71.51& 67.45 & 75.28 \\

$\text{S}^{2}\text{A-Net}$ \cite{han2021align} & R-101 & & 88.70 & 81.41 & 54.28 & 59.75 & 78.04 & 80.54 & 88.04 & 90.69 & 84.75 & 86.22 & 65.03 & 65.81 &76.16 & 73.37 & 58.86 & 76.11 \\
PolarDet \cite{zhao2021polardet} & R-101 & $\checkmark$ & 89.65& 87.07& 48.14& 70.97& 78.53& 80.34& 87.45& 90.76& 85.63& 86.87& 61.64& 70.32& 71.92& 73.09& 67.15 & 76.64 \\

DAL ($\text{S}^{2}\text{A-Net}$) \cite{ming2021dynamic} & R-50& $\checkmark$&  89.69 &83.11 &55.03& 71.00& 78.30& 81.90 &88.46 &90.89& 84.97 & 87.46 &64.41 &65.65& 76.86 &72.09& 64.35& 76.95 \\

GGHL \cite{GGHL} &D-53 & & 89.74& 85.63& 44.50& 77.48& 76.72& 80.45& 86.16& 90.83& 88.18& 86.25& 67.07& 69.40& 73.38& 68.45& 70.14&  76.95 \\

DCL ($\text{R}^{3}\text{Det}$) \cite{yang2021dense} & R-152 & $\checkmark$ & 89.26 &83.60 &53.54 &72.76 &79.04 &82.56 &87.31 &90.67 &86.59 &86.98 &67.49 &66.88 &73.29 &70.56 &69.99 & 77.37  \\

RIDet \cite{ming2021optimization} & R-50 & & 89.31 &80.77 &54.07 &76.38 &79.81 &81.99 &\textbf{89.13} &90.72& 83.58& 87.22& 64.42 &67.56 &\textbf{78.08}& 79.17& 62.07& 77.62 \\
\hline

QBB (baseline)  & R-50 & & 77.52&57.38&37.20  &65.97 &56.29 &69.99 &70.04 &90.31 &81.14 &55.34 &57.98 &49.88 &56.01 &62.32&58.37&63.05 \\
PointSet (baseline)  & R-50 & & 87.48 & 82.53&45.07&65.16&78.12&58.72&75.44&90.78&82.54&85.98&60.77&67.68&60.93&70.36 &44.41 & 70.39 \\
\textbf{G-Rep} (QBB)  & R-101 & & 88.89&74.62 &43.92 &70.24 &67.26 &67.26 &79.80 &90.87 &84.46 &78.47 &54.59 &62.60 &66.67 &67.98 &52.16 &70.59   \\ 
\textbf{G-Rep} (PointSet)  & R-50 & & 87.76 &81.29 & 52.64 & 70.53 & 80.34 & 80.56 & 87.47 & 90.74 &82.91 & 85.01 &61.48 & 68.51 & 67.53 & 73.02 &63.54 & 75.56 \\ 

\textbf{G-Rep} (PointSet)  & RX-101 & $\checkmark$ & 88.98 &79.21 & 57.57 & 74.35 & \textbf{81.30} & 85.23 &  88.30 & 90.69 &85.38 & 85.25 &63.65 & 68.82 & 77.87 & 78.76 &71.74 & 78.47 \\ 
\textbf{G-Rep}(PointSet) & Swin-T & $\checkmark$ & 88.15 & 81.64 & \textbf{61.30} & 79.50 & 80.94 & \textbf{85.68} & 88.37 & \textbf{90.90} & 85.47 & \textbf{87.77} & \textbf{71.01} & 67.42 & 77.19 & \textbf{81.23} & \textbf{75.83} & \textbf{80.16}\\ 
\hline
\end{tabular}}
\end{table}

\subsection{Comparison with other Methods}
The performance of the proposed method is compared with that of other state-of-the-art detection methods on the DOTA dataset, which is a benchmark aerial image dataset for multi-class oriented object detection. 
The experimental results are shown in \tabref{tab:DOTARes}, where R-50/101, RX-101, D-53, H-104 and Swin-T denote ResNet-50/101 \cite{he2016deep}, ResNeXt-101 \cite{xie2017aggregated}, DarkNet53 \cite{redmon2018yolov3}, Hourglass-104 \cite{newell2016stacked}, and Swin-Transformer \cite{liu2021swin}, respectively.
The results in \tabref{tab:DOTARes} show that the one-stage methods exhibits a trend of outperforming the two-stage methods, and the G-Rep based on anchor-free method RepPoints achieves the state-of-the-art performance.
With the excellent backbone Swin-T \cite{liu2021swin}  and some common tricks, such as data augmentation, G-Rep achieves 80.16\% mAP on the DOTA dataset.
In addition, results show that G-Rep performs better on some object categories with large aspect ratios (e.g., BR, LV, and HC).
Based on the anchor-free baseline, G-Rep is superior to the anchor-free method ACE \cite{dai2022ace} 8.64\% mAP.
G-Rep is a pioneering new paradigm in oriented object detection.

\begin{table}[htb]
\caption{Comparison of the mAP of various rotation methods on HRSC2016.} \label{tab:HRSCRes}
\begin{center}\footnotesize
\begin{tabular}{c|c}
\hline
\textbf{Method}& \textbf{mAP(\%)} \\ \hline
RoI-Transformer \cite{ding2018learning}& 86.20 \\
RSDet \cite{qian2021learning}& 86.50 \\
Gliding Vertex \cite{xu2020gliding} & 88.20 \\
BBAVectors \cite{yi2021oriented} & 88.60 \\
$ \text{R}^{3}\text{Det}$ \cite{yang2021r3det} &89.26 \\
DCL \cite{yang2021dense} &\textbf{89.46} \\
\hline
\textbf{G-Rep} (QBB) & 88.02 \\
\textbf{G-Rep} (PointSet) & \textbf{89.46} \\
\hline
\end{tabular}
\end{center}
\end{table}

\begin{table}[htb]
\caption{Comparison of the AP with state-of-the-art methods on UCAS-AOD.} \label{tab:UCASRes}
\begin{center}\footnotesize
\begin{tabular}{c|cc|c}
\hline
\textbf{Method}& \textbf{car} & \textbf{airplane}  &\textbf{mAP(\%)} \\
\hline
RetinaNet \cite{lin2017focal} &84.64 &90.51 &87.57 \\
Faster-RCNN \cite{ren2015faster}  &86.87&89.86&88.36\\
RoI-Transformer \cite{ding2018learning}  & 88.02&90.02&89.02 \\
RIDet-Q \cite{ming2021optimization} &88.50&89.96&89.23 \\
RIDet-O \cite{ming2021optimization} &88.88&90.35&89.62 \\
DAL \cite{ming2021dynamic} & 89.25&90.49&89.87 \\
\hline
 \textbf{G-Rep} (QBB)  & 87.35 & 90.30& 88.82 \\
 \textbf{G-Rep} (PointSet) & \textbf{89.64} & \textbf{90.67} & \textbf{90.16}  \\ 
\hline
\end{tabular}
\end{center}
\end{table}

To further verify the effectiveness of the proposed method and compare it with other state-of-the-art detectors, a series of experiments were conducted on HRSC2016 and UCAS-AOD, and the results are shown in \tabref{tab:HRSCRes} and \tabref{tab:UCASRes}, respectively.
The results demonstrate that the models using G-Rep achieve state-of-the-art performance.
Although DCL \cite{yang2021dense} achieves the same performance as G-Rep (PointSet) on the ship detection dataset HRSC2016, G-Rep does not require preset anchors and requires fewer hyper-parameters and computations, proving the robustness of G-Rep.
\tabref{tab:UCASRes} shows that G-Rep (PointSet) gets 0.93\% and 0.54\% higher mAP than RIDet (OBB) and RIDet (QBB) \cite{ming2021optimization}, respectively, indicating the superiority of using Gaussian representation for objects.

To further demonstrate the generalizability of G-Rep, we also conducted some extended experiments on the oriented text detection dataset ICDAR2015. The experimental results are listed in \tabref{tab:ICDARRes}.
Precision, Recall, and F-measure were used as the evaluation metrics that follow the official criteria of the ICDAR2015 dataset.
The experimental results show that G-Rep (PointSet) outperforms the one-stage method RO$^{3}$D \cite{hou2022refined} and the two-stage method SCRDet \cite{yang2019scrdet} on Recall and F-measure, demonstrating the universality and superior performance of G-Rep.

\begin{table}[htb]
\caption{Comparison of the performance of different methods on ICDAR2015.} \label{tab:ICDARRes}
\begin{center}\footnotesize
\begin{tabular}{c|ccc}
\hline
\textbf{Method}&  \textbf{Precision} & \textbf{Recall}& \textbf{F-measure}\\ \hline
GWD \cite{yang2021rethinking} & 80.5&  74.0&  77.1 \\
RRPN \cite{ma2018arbitrary} & 82.2 & 73.2 & 77.4 \\
SCRDet \cite{yang2019scrdet} & 81.3 & 78.9 & 80.1 \\
RO$^{3}$D \cite{hou2022refined}  &\textbf{83.9}& 78.6 & 81.2\\ 
\hline 
\textbf{G-Rep} (QBB) & 80.5  & 71.4 &  75.8 \\
\textbf{G-Rep} (PointSet) & 81.6 & \textbf{81.1} & \textbf{81.3}  \\ 
\hline
\end{tabular}
\end{center}
\end{table}

\subsection{Visualization Analysis}
\begin{figure}[htb]
\centering
\includegraphics[width=0.99\linewidth]{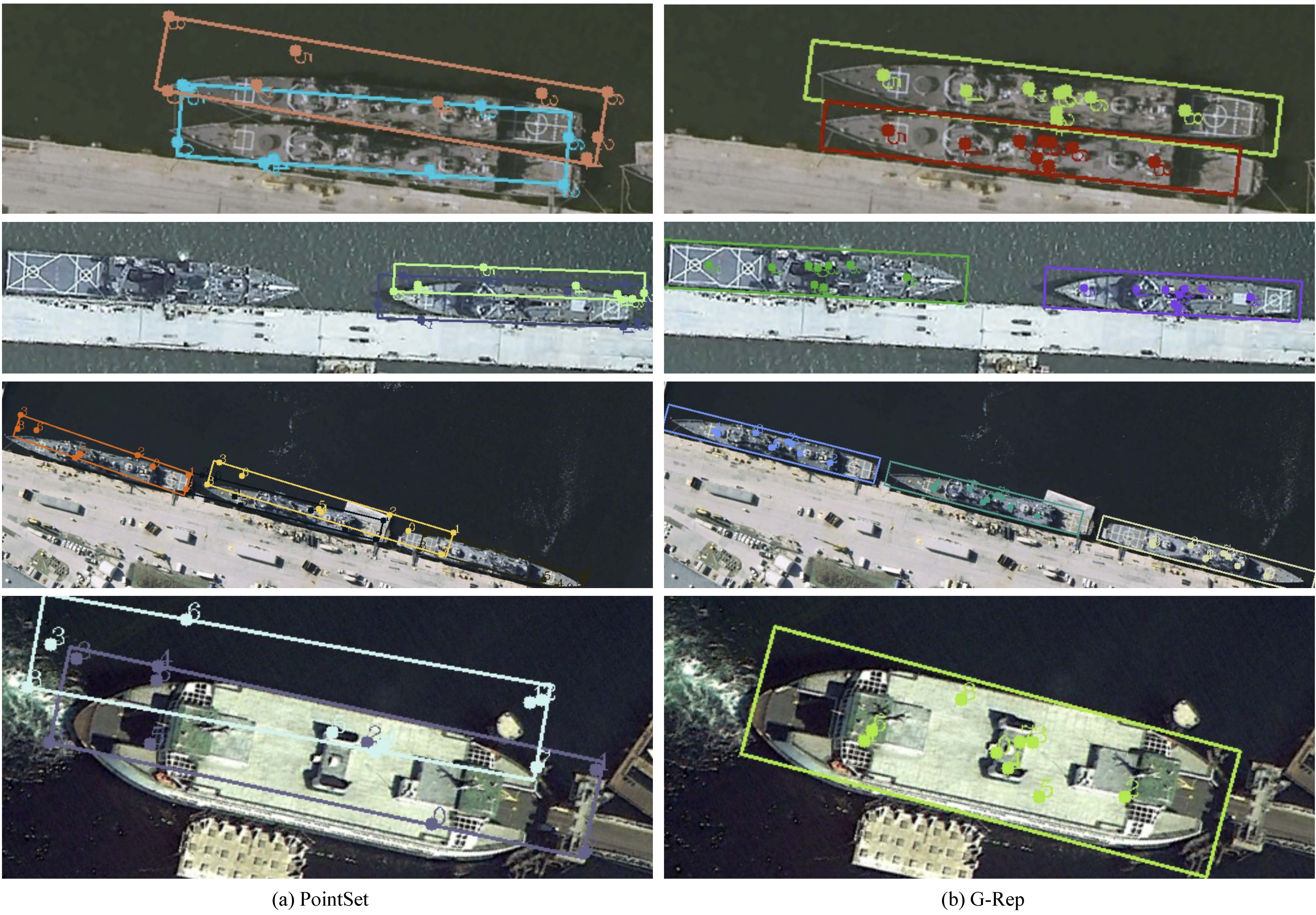}
\caption{Comparison of the visualization results of PointSet and G-Rep on HRSC2016 dataset.} \label{fig:VisCompareResults_HRSC2016}
\end{figure}

\begin{figure}[htb]
\centering
\includegraphics[width=0.75\linewidth]{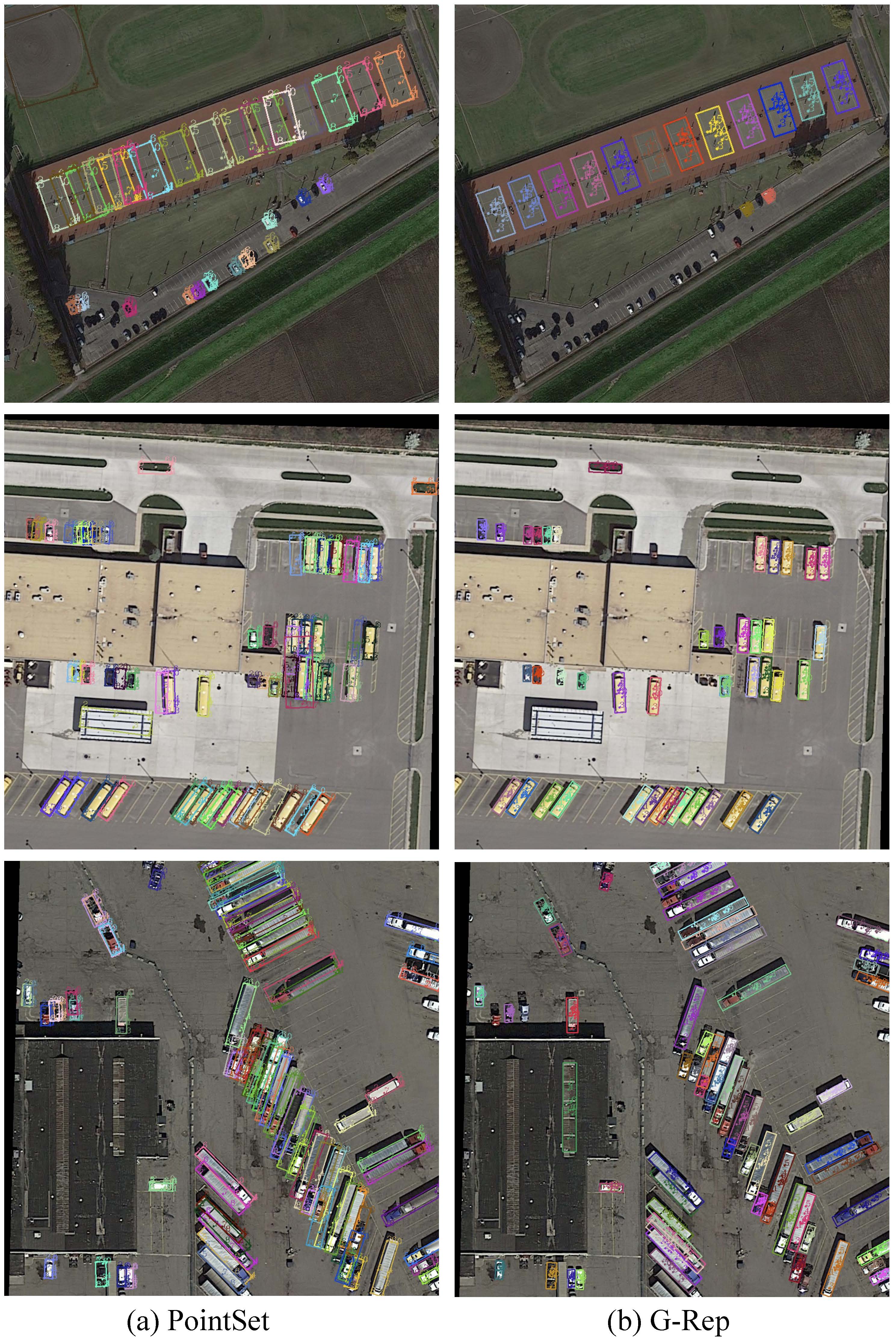}
\caption{Comparison of the visualization results of PointSet and G-Rep on DOTA dataset.} \label{fig:VisCompareResults_DOTA}
\end{figure}

\begin{figure}[htb]
\centering
\includegraphics[width=0.8\linewidth]{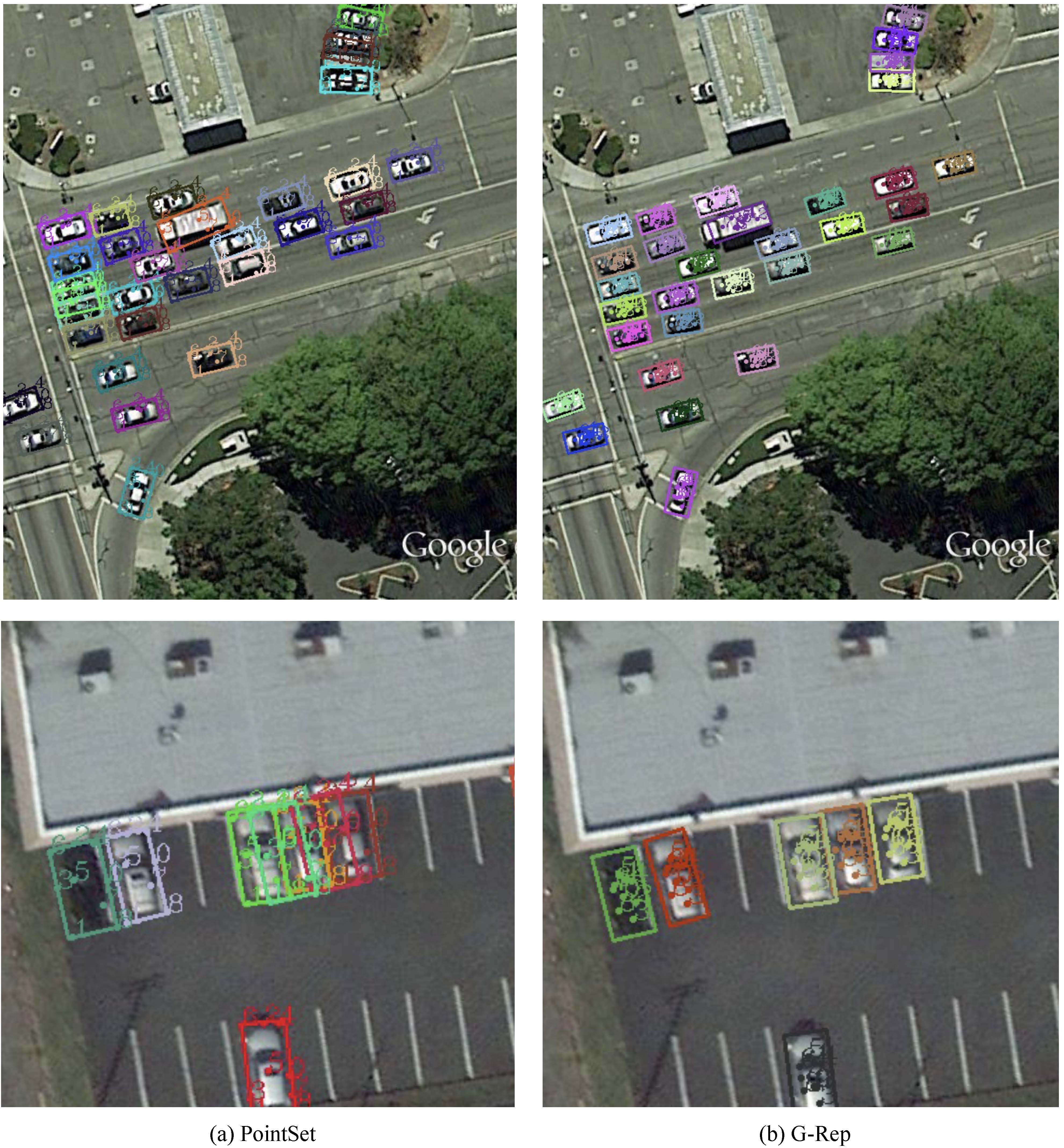}
\caption{Comparison of the visualization results of PointSet and G-Rep on UCAS-AOD dataset.}\label{fig:VisCompareResults_UCASAOD}
\end{figure}

\figref{fig:VisCompareResults_HRSC2016} visualizes the detection results of PointSet and G-Rep on HRSC2016. The points of PointSet are distributed at the boundary of objects, while the points of the G-Rep are distributed in the interior of the object.
Therefore, G-Rep is superior in terms of accurate localization and is not sensitive to outliers.
More visualization examples of the experimental results on DOTA and UCAS-AOD datasets are shown in \figref{fig:VisCompareResults_DOTA} and \figref{fig:VisCompareResults_UCASAOD}.
The visualization results show that points of G-Rep are concentrated in the interior of the object, which is a relatively accurate localization of the object. The points of PointSet, on the other hand, focus on the boundaries of the object, which can easily cause deviations in localization.
Additionally, these visualization results of different kinds of objects on different datasets sufficiently demonstrate the superiority of G-Rep.

\clearpage

\subsection{Discussion of Limitation}
Although the proposed method provides a uniform representation of the Gaussian distribution for various representations of the input, an obvious limitation is that the format of the output can only be OBB. The output points are dispersed as the Gaussian distribution inside the object, and OBB is transformed from the Gaussian distribution of the output points. Additionally, the angle prediction of the square-like object transformed by the isotropic Gaussian is inaccurate.
Note that the irregular PointSet and QBB representations are more accurate than the OBB representation for most objects. However, for some square-like regular objects, OBB representation is more accurate than PointSet and QBB representations.

\section{Conclusions} \label{section:Conclusions}
The main contribution of this study is that G-Rep is proposed to construct the Gaussian distribution on PointSet and QBB, which overcame the limitation of Gaussian applications for current object detection methods and truly achieved a unified solution.
Thus, a series of detection challenges resulted from object representations are alleviated.
Additionally, various label assignment strategies for the Gaussian distribution are designed, and the metrics are aligned between label assignment and regression loss.
More importantly, G-Rep overcomes the current dilemma, and provides inspiration for exploring other forms of label assignment strategies and regression loss functions.

\section*{Acknowledgements}
	This work was supported in part by the National Natural Science Foundation of China (61731022, 61871258, 61929104, U21B2049), the NSFC Key Projects of International (Regional) Cooperation and Exchanges (61860206004) and the Key Project of Education Commission of Beijing Municipal (KZ201911417048). 

\bibliographystyle{elsarticle-num}
\bibliography{G-Rep-v2}

\end{document}